\documentclass[runningheads]{llncs}

\usepackage{eccv}

\usepackage{eccvabbrv}

\usepackage{hyperref}

\usepackage{url}            
\usepackage{booktabs}       
\usepackage{nicefrac}       
\usepackage{microtype}      
\usepackage{xcolor}         
\usepackage{fleqn, tabularx}
\usepackage{multirow}
\usepackage{mathtools}
\usepackage[export]{adjustbox}
\usepackage{colortbl}
\usepackage{wrapfig}
\usepackage{algorithm}
\usepackage[noend]{algorithmic}
\usepackage{xcolor}
\usepackage{enumitem}
\usepackage[section]{placeins}
\usepackage{orcidlink}
\usepackage{graphicx}

\renewcommand{\hat}{\widehat}

\begin{document}
\title{Hierarchical B-frame Video Coding \\ for Long Group of Pictures}

\author{Kirillov Ivan$^*$ \and
Parkhomenko Denis$^*$ \and
Chernyshev Kirill$^*$ \and
Pletnev Alexander$^*$ \and
Yibo Shi \and
Kai Lin$^*$ \and
Babin Dmitry}

\authorrunning{I.~Kirillov et al.}

\institute{Huawei Technologies}

\def\thefootnote{*}\footnotetext{These authors contributed equally to this work}

\maketitle

\begin{abstract}
Learned video compression methods already outperform VVC in the low-delay (LD) case, but the random-access (RA) scenario remains challenging.
Most works on learned RA video compression either use HEVC as an anchor or compare it to VVC in specific test conditions,
using RGB-PSNR metric instead of Y-PSNR and avoiding comprehensive evaluation.
Here, we present an end-to-end learned video codec for random access that combines training on long sequences of frames,
rate allocation designed for hierarchical coding and content adaptation on inference.
We show that under common test conditions (JVET-CTC), it achieves results comparable to VTM (VVC reference software)
in terms of YUV-PSNR BD-Rate on some classes of videos, and outperforms it on almost all test sets in terms of VMAF BD-Rate.
On average it surpasses open LD and RA end-to-end solutions in terms of VMAF and YUV BD-Rates.
\end{abstract}

\section{Introduction}
\label{sec:introduction}

Video compression has long been a critical aspect of digital media, enabling the efficient storage and transmission of video content. 
It plays a significant role in applications such as video streaming and video conferencing, 
which require efficient compression techniques to effectively transmit data over limited bandwidth networks. 
Among the many video coding methods, two prominent approaches stand out: low-delay P-frame coding and random access coding.

\begin{figure}[tb]
    \centering
    \includegraphics[width=0.7\linewidth]{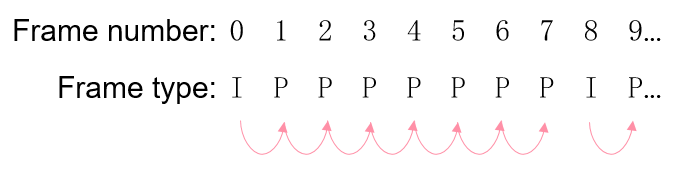}
    \caption{Low-delay coding scheme when \textit{group of pictures} (GoP) length equals 8. 
    Arrows represent reference structure: each P-frame refers to the previous frame.}
    \label{fig:ldp_gop}
\end{figure}

\textbf{Low-delay P-frame} coding prioritizes minimizing the time delay between encoding and decoding to facilitate real-time applications such as videoconferencing and live streaming. 
In this approach, frames are typically coded in a predictive manner, where each frame is predicted based on the previous frame (see \cref{fig:ldp_gop}).

\textbf{Random access} coding emphasizes the ability to access any frame independently, 
without the need to fully decode preceding or succeeding frames. 
The key component of the RA coding scheme is a bidirectional frame (B-frame).
B-frames use both past and future reference frames for prediction (see \cref{fig:gop-a}), resulting in higher compression efficiency compared to P-frames.

While both low-delay P-frame coding and random access coding have their respective strengths and weaknesses, 
the choice between them depends on the specific requirements of the application, 
with considerations such as latency, error resilience, and compression efficiency playing key roles in the decision-making process.

The current standard in this area are traditional codecs such as HEVC \cite{hevc} and VVC \cite{vvc}.
They use established algorithms to reduce file size while maintaining visual quality. 
However, the emergence of neural network based hybrid and end-to-end codecs represents a significant advancement in the field.

The traditional RA coding scheme (see \cref{sec:gopstructure}) assumes that different B-frames have different distances to the references.
The main problem for end-to-end RA coding is the high variability of motion between reference and target frames.
For data-centric machine learning approaches, this is a challenging task due to the complex distribution of the input data.
Some papers, such as \cite{IBVC}, propose to use a different coding scheme to solve this problem.
Such an approach makes the distance to the references constant, thus reducing motion variability. 
However, it is not scalable and less consistent with the random access ideology, 
since decoding a new frame requires on average several times more already decoded frames.
Here, we follow the traditional hierarchical coding scheme and focus on improving the generalization ability of the B-frame model, 
by training it to perform equally well with references at different distances.

In this work, we present an end-to-end solution for video compression in random access case, which on average outperforms all open end-to-end solutions.
We present a novel training approach for the RA scenario, which allowed us to achieve a noticeable performance gain in terms of BD-rate \cite{bdrate}.
This approach involves the use of longer training sequences, a special data sampling technique, and a loss function (see \cref{sec:training}).
The model works with a commonly used hierarchical coding scheme (see \cref{sec:gopstructure})
and we describe architectural changes introduced for such a scheme (see \cref{sec:bframe}).
One of these changes is a novel Hierarchical Gain Unit block, added to incorporate hierarchy level-adaptive latent scaling (see \cref{sec:hgu}).
Finally, we describe the RA-oriented content adaptation technique (see \cref{sec:contad}).

\section{Related Work}
\label{sec:related_work}

\subsection{Learned Image Compression}
Neural network based image compression has shown great results in the last years, achieving and overcoming the level of traditional codecs.
First notable results were demonstrated in the papers of Balle \etal \cite{Balle2017, Balle2018}, 
where variational auto-encoder architecture was unified with factorized and hyperprior entropy models.
Later, challenging H.266 (VTM) intra-frame coding, the Gaussian mixture model \cite{Cheng2020} was introduced. 
For variable bitrate image coding within one model, Gain Units\cite{GainUnit} were proposed.

\subsection{Learned Video Compression}
DVC\cite{DVC} was the first work to propose an end-to-end video codec. 
It could only compete with H.264, but it's core ideas laid the foundation for further research. 
Taking from DVC decomposition on motion and residual, FVC\cite{FVC} performed alignment, 
known in traditional codecs as motion compensation, in latent space. 
Among other innovations, it helped to outperform H.265 on popular benchmarks. 
Later, AlphaVC\cite{AlphaVC} took this approach, built its P-frame and conditional I-frame on it, 
and reported impressive gains over VVC in objective metrics such as PSNR in YUV color space.
Another line of research can be attributed to the DCVC series of papers \cite{DCVC, DCVCTCM, DCVCHEM, DCVCDC,DCVCFM}, 
where the superiority of conditional coding over residual coding was claimed and experimentally proven.

The success of these models, despite the limitations of the low-delay scenario, may explain why some learned random access methods use them. 
IBVC \cite{IBVC} propose coding scheme mixing novel B-frames, with interpolated image as context, and P-frames from \cite{DCVCHEM}. 
Besides the dependence on the low-delay model, such coding scheme doesn't fully exploit the temporal redundancy, which affects the overall performance.
Hi-DCVC \cite{HiDCVC} follows hierarchical coding and modified DCVC architecture to work with two references. 
Trade-offs between quality and allocated bitrate between frames on different temporal layers are controlled by loss function and latent scaling.

However, results in all these papers are computed in RGB-PSNR, even though traditional codecs prefer YUV color space.

\section{Method}
\label{sec:approach}

\subsection{Overview}

\begin{figure}[tb]
    \centering
    \begin{subfigure}{0.48\linewidth}
        \includegraphics[width=\linewidth]{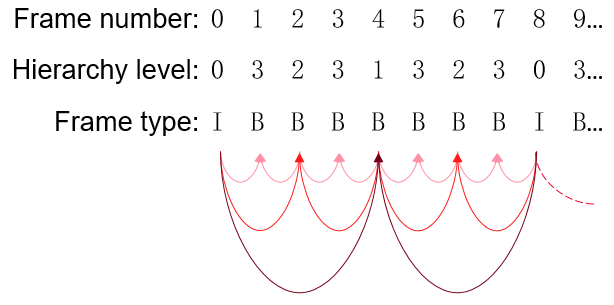}
      \caption{}
      \label{fig:gop-a}
    \end{subfigure}
    \hfill
    \begin{subfigure}{0.48\linewidth}
        \includegraphics[width=\linewidth]{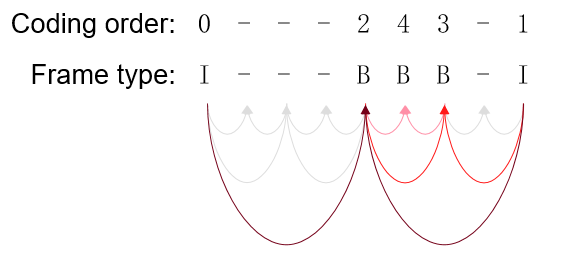}
      \caption{}
      \label{fig:gop-b}
    \end{subfigure}
    \caption{ \textbf{(a)}:GoP structure for the hierarchical coding scheme when the GoP size is 8. 
    Arrows connect frames to their child frames (which use them as reference). 
    Coding order may vary, but should always satisfy the condition that frame $x_t$ is coded before its child frames. 
    \textbf{(b)}:\textit{Random Path} sampling strategy when the training GoP size is 8. 
    Only frames with colored incoming/outgoing arrows are compressed.}
    \label{fig:gop}
  \end{figure}

A video sequence consists of ordered frames, and during encoding/decoding they are divided into groups of pictures. 
The most common approach in random access scenario is hierarchical coding (\cref{fig:gop-a}).
The first and last frames of each GoP are coded as intra-frames (I-frames), while others are coded as B-frames.

\begin{figure}[tb]
    \centering
    \includegraphics[width=1.0\linewidth]{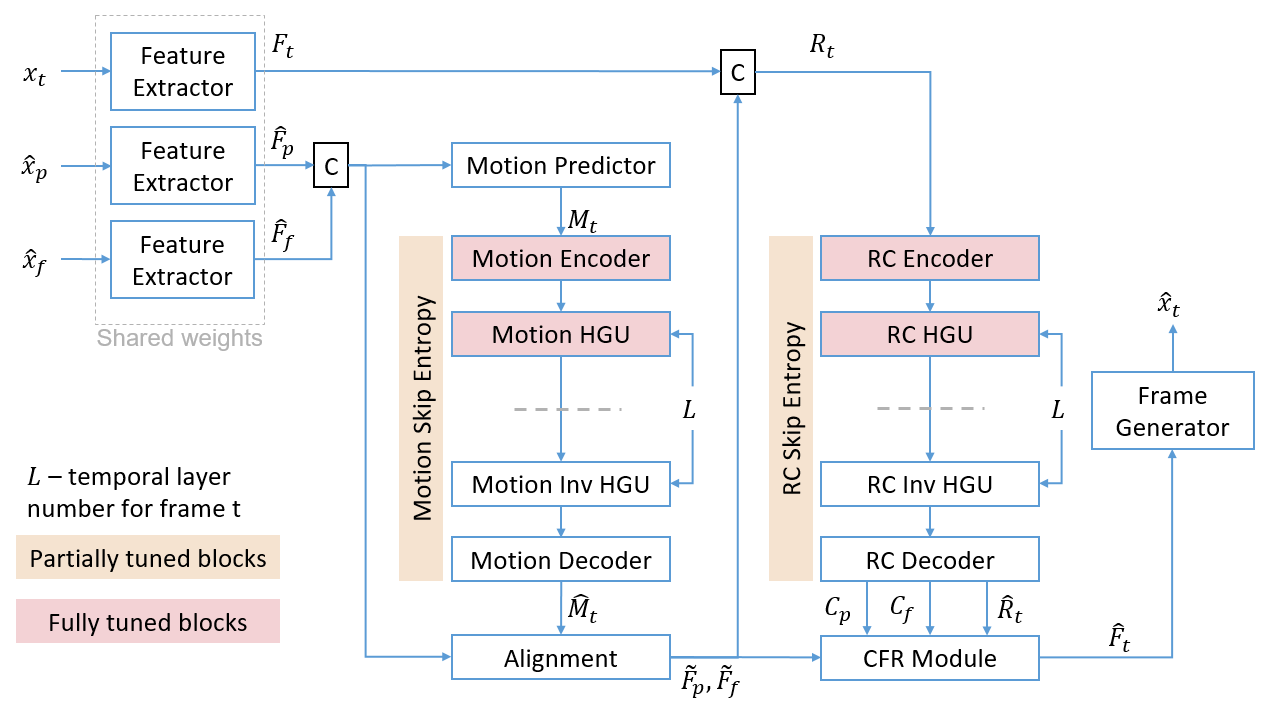}
    \caption{Overview of the proposed B-frame compression model.}
    \label{fig:codec-scheme}
\end{figure}

The architecture of the proposed B-frame is shown on \cref{fig:codec-scheme}.
It is inspired by the HVFVC \cite{hvfvc} P-frame model, 
extending it for the RA case and especially for the hierarchical coding scheme.

Next, we will discuss in detail the major changes made to the P-frame approach: 
shift to the hierarchical coding GoP structure (in architecture and training), 
architectural modifications of the model for two reference frames, 
content adaptation procedure for hierarchical coding,
training optimization and loss selection.

\subsection{Hierarchical Coding}
\label{sec:gopstructure}

The hierarchical coding scheme may be formalized as follows. Denote a video sequence as $X=\{x_0 , x_1 , ...\}$. 
Suppose that the GoP size is set to $GoP=2^n, n \in \mathbf N$.
In this case, frames $x_{GoP \cdot i}, i=0, 1, ...$ are coded as I-frames and others coded as B-frames.
If we already have two decoded frames $\hat x _p, \hat x _f$ such that $f - p = 2\Delta _t \geq 2$ and no decoded frames $x_k, p<k<f$,
then we can use them as reference frames to encode and decode frame $x_t$:

\begin{align}
&t = p + \Delta _t = f - \Delta _t ,\\\nonumber
&\hat{x}_t = Dec(Enc(x_t, \hat x _p, \hat x _f), \hat x _p, \hat x _f),\\\nonumber
&\text{where } Enc \text{ is the encoder part of the model and } Dec \text{ is the decoder part}.\nonumber
\end{align}

After that we will have two pairs of decoded frames $(\hat x _p, \hat x _t)$ and $(\hat x _t, \hat x _f)$ and can repeat the scheme recursively.
We denote the hierarchy level of the frame $t$ by $level(t) = log _2 \left(\frac{GoP}{\Delta _t}\right)$.
In the proposed approach, this scheme is used not only during testing, but also during training (see \cref{sec:training}).

\subsection{B-frame model}
\label{sec:bframe}

As it was mentioned, the proposed B-frame model is inspired by HVFVC \cite{hvfvc} P-frame model, 
so our architecture uses the same key modules and extends some of them to two-references case. 
We also introduce new Hierarchical Gain Unit (HGU) module, which will be discussed in \cref{sec:hgu}.

The B-frame encoder takes three inputs: current frame $x_t$, and two reconstructed reference frames $\hat x _p , \hat x _f$. 
The Feature Extractor module transforms each input into corresponding features $F_t , \hat F _p , \hat F _f$, 
reducing the spatial dimensions by a factor of 2.

\begin{figure}[tb]
    \centering
    \begin{subfigure}{0.53\linewidth}
        \includegraphics[width=\linewidth]{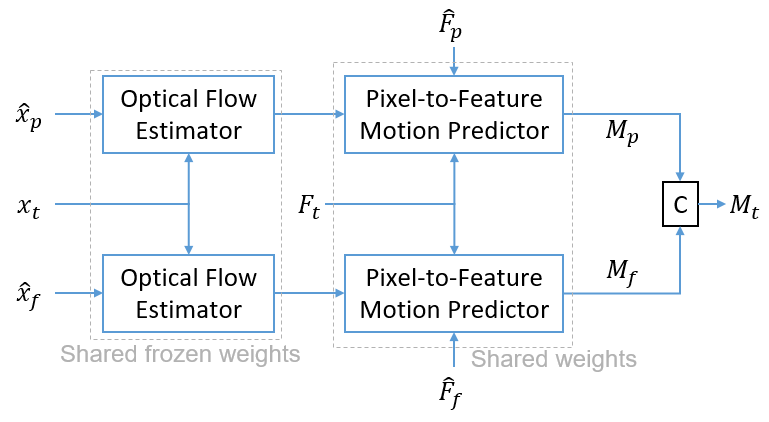}
        \caption{}
        \label{fig:motion_predictor}
    \end{subfigure}
    \hfill
    \begin{subfigure}{0.43\linewidth}
        \includegraphics[width=\linewidth]{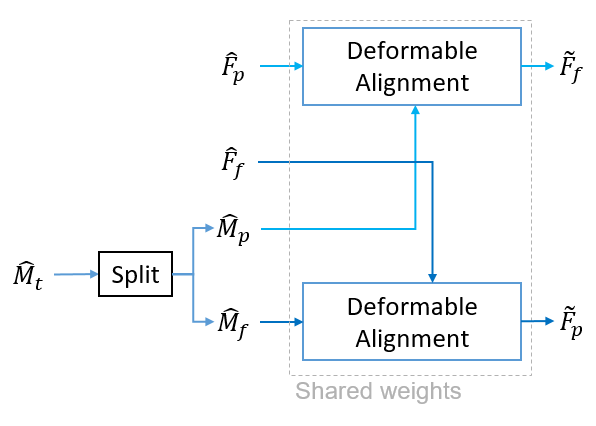}
        \caption{}
        \label{fig:alignment}
    \end{subfigure}
    \caption{\textbf{(a)}: B-frame Motion Predictor scheme. First, the motion flows $OFE(\hat x _p , x_t)$ and $OFE(\hat x _f , x_t)$ 
    are predicted using the pretrained Optical Flow Estimator $OFE$.
    These flows are used as an initialization for the deformable alignment offsets within the Pixel-to-Feature Motion Predictor module.
    It generates motion features $M _p, M _f$ from the $F_t$, corresponding flows, and reference features.
    \textbf{(b)}: Alignment module scheme. Deformable alignment offsets are initialized with the reconstructed motion features $\hat M _p , \hat M _f$ 
    and applied to the corresponding reconstructed features $\hat F _p , \hat F _f$.} 
    \label{fig:motion_predictor_alignment}
  \end{figure}

Motion Predictor (\cref{fig:motion_predictor}) predicts two motion features $M_p, M_f$ and returns their concatenation $M_t$. 
Both features are predicted using the pixel-to-feature motion prediction method proposed in \cite{AlphaVC}. 
In this way, the initial flow predicted by the pretrained optical flow estimator is converted into motion in the feature domain, 
and additionally refined with convolutional layers.
We use the LiteFlowNet \cite{liteflownet} as a flow estimator in the early stages of training for efficiency
and replace it with the more accurate but slower RAFT \cite{raft} in the last stage.
This way we save up to 30\% of training time with no loss of quality on the inference.

After that $M_t$ is passed through the Motion Encoder and the new Motion HGU module (\cref{fig:hgu_scheme}). 
The Motion HGU scales the motion latent by a value depending on the hierarchy level of the current frame (see \cref{sec:hgu}). 

Then, motion information is transmitted to the decoder side using Motion Skip Entropy. 
This module utilizes the efficient probability-based entropy skipping method, introduced in \cite{AlphaVC}.

On the decoder side, the bitstream is decoded and passed through the Motion Inv HGU and Motion Decoder, resulting in a decoded motion feature $\hat M _t$.

After that we align the features $\hat F _p , \hat F _f$ to get predictions $\tilde F _p , \tilde F _f$. 
Inside the Alignment module (\cref{fig:alignment}) we split $\hat M _t$ into $\hat M _p , \hat M _f$ along the channel dimension, 
then pass the pairs $\hat F _p , \hat M _p$ and $\hat F _f , \hat M _f$ through the Deformable Alignment block.
This way, instead of generating one feature prediction or one conditioning feature, we choose to utilize two separate predictions: 
one from the past and one from the future.
Feature merging leads to additional loss of predicted information, resulting in less efficient signal compression.
Therefore, we choose to merge two predicted features into one only in the Confidence-based Feature Reconstruction (CFR) module,
after residual and confidence (RC) maps encoding and decoding.

\begin{figure}[tb]
    \centering
    \includegraphics[width=0.6\linewidth]{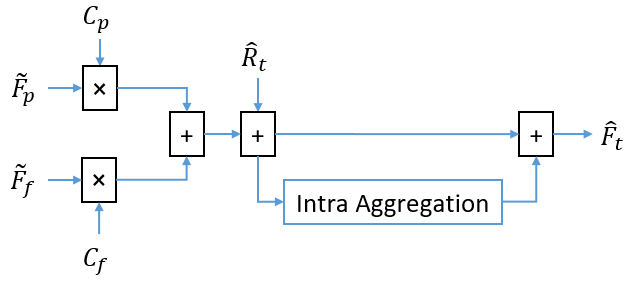}
    \caption{CFR module scheme. 
    First, feature $\check F _t$ is calculated as $\check F _t = C_p \cdot \tilde F _p + C_f \cdot \tilde F _f + \hat R _t$.
    Then it is refined using the Intra Aggregation block, producing $\hat F _t$}
    \label{fig:cfr}
\end{figure}

Predictions $\tilde F _p , \tilde F _f$ are concatenated together with $F_t$ and passed to RC Encoder and RC HGU modules.
The RC Skip Entropy module is used to transmit information to the decoder side, where it passes through RC Inv HGU and RC Decoder.
Following HVFVC, we utilize confidence-based feature reconstruction method, extending it to the two references case.
Thus, the RC Decoder predicts two confidence maps $C_p , C_f$ for corresponding predictions  $\tilde F _p , \tilde F _f$, and the residual $\hat R _t$. 
Along with the predictions these values are passed to the CFR Module (\cref{fig:cfr}) 
which computes the reconstructed feature $\hat F _t$.
Details on CFR and Intra Aggregation modules can be found in \cite{hvfvc}.
Finally, the Frame Generator transforms $\hat F _t$ into the reconstructed frame $\hat x _t$.

Confidence-based feature reconstruction was chosen because of its flexibility in the use of reference features.
In the case of a hierarchical coding scheme, reference frames at different distances will contain different numbers of hardly predictable regions 
such as occlusions and newly emerged objects. 
In other words, close references will have more information in common with the target frame than distant references.
Confidence maps will exclude useless regions from the predicted features.

\subsection{Hierarchical Gain Unit}
\label{sec:hgu}

To incorporate knowledge about the hierarchical coding structure into our model, we propose to add level-dependent scaling of latents.
In particular, we adopted Gain Units \cite{GainUnit} to learn specific scaling parameters 
for motion and residual latents at each level of the coding hierarchy.
The idea is that for frames with large distances to the references, we need to spend more bits to get proper reconstruction quality. 
Additional level-adaptive scaling leads to an adjustment of the number of unique values in the quantized latent, 
which affects both the bitrate and the reconstruction quality.

First, the latent is scaled on the encoder side:

\begin{align}
l_t^q[c][h][w] = l_t[c][h][w] \cdot q_{enc}^{level(t)}[c]
\end{align}

where $l_t$ -- motion or RC latent, $q_{enc}^k$ -- trainable scaling parameter for level $k$ on encoder. 

And then the invert scaling is applied on the decoder side:

\begin{align}
\hat{l}_t[c][h][w] = \hat{l_t^q}[c][h][w] \cdot q_{dec}^{level(t)}[c]
\end{align}

where $\hat{l_t^q}$ -- reconstructed motion or RC latent, $q_{dec}^k$ -- trainable scaling parameter for level $k$ on decoder. 
Scaling and inverse scaling parameters can be $c$-dimensional for channel-wise scaling of the latent with $c$ channels 
or scalar for scaling the entire tensor. 
Our experiments didn't show any significant difference between scalar and vector approaches, so the scalar approach was chosen as the simpler one.

\begin{figure}[tb]
  \centering
    \begin{subfigure}{0.58\linewidth}
        \includegraphics[width=\linewidth]{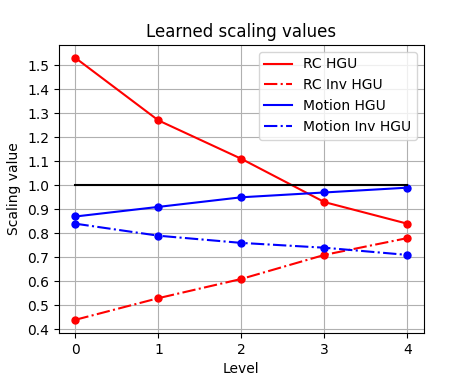}
        \caption{}
        \label{fig:hgu_values}
    \end{subfigure}
    \hfill
    \centering
    \begin{subfigure}{0.28\linewidth}
        \includegraphics[width=\linewidth]{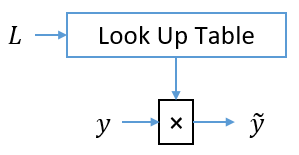}
        \caption{}
        \label{fig:hgu_scheme}
    \end{subfigure}
    \caption{\textbf{(a)}: Level-adaptive scaling values, learned by our model. 
    \textbf{(b)}: Hierarchical Gain Unit scheme.
    Scaling value $q$ is taken from the Look Up table by the index of current hierarchy level $L$.
    }
    \label{fig:hgu}
  \end{figure}

\cref{fig:hgu_values} shows the actual scaling values learned by our model. 
Interestingly, the level-adaptive scaling behaves differently for motion and RC latents.
The RC scaling values decrease as the hierarchy level increases. 
This means that RC latents for frames with small reference distances $\Delta _t$ contain less information 
(after quantization) than those with larger $\Delta _t$. 
On the contrary, motion latent scaling values increase as the hierarchy level increases.
This is probably due to motion precision and amplitude at different distances.
Accurately estimated motion between close frames requires fine granularity, while high-amplitude motion could be coarsely quantized.

Scaling coefficients missing in training are approximated on inference by an exponential function. 
Therefore, training strategy with large GoP size, which this work follows (see \cref{sec:training}), reduces the uncertainty in the scaling parameter, 
leading to more efficient bit allocation.

\subsection{Content Adaptation}
\label{sec:contad}

Following previous work on low-delay neural compression \cite{caldp}, 
our method also optimizes the rate distortion function on test frames and thus adapts to the sequence. 
Such online training doesn't change the decoder weights and therefore doesn't require additional bits to be transmitted. 
Due to the variation of the reference distance, the B-frame benefits from the tuning of motion-related modules.
As for RC modules, our experiments showed that their optimization is harmful to overall performance, but can be used on non-reference ($t \mod 2 = 1$) frames. 
Modules tuned during content adaptation are marked with a color on \cref{fig:codec-scheme}.
Entropy modules are marked as partially tuned because only encoder-side submodules are tuned in our setup. 

It was observed that content adaptation may be beneficial not only for B-frames but also for I-frames. 
Moreover, for sequences with little motion, only I-frame online training brings reasonable gain. 
Content adaptation leaves freedom in the choice of loss function and is not limited to PSNR as distortion.
For example, the subjective quality of the compressed video can be improved by tuning modules with perceptual metrics, 
such as LPIPS\cite{LPIPS}.

Frames are tuned independently in their decoding order with RD loss ($\lambda$ as in training, MSE as distortion), 
taking into account only the current frame.

\subsection{Training}
\label{sec:training}

Common test conditions for the RA scenario \cite{jvetctc} imply that the GoP size is usually 32 or 64 frames.
This means that the distance to the reference frames $\Delta _t$ can be as large as 32 frames, 
which leads to strong differences between the reference frames and the current frame.
To address this problem, we propose the following: 
different to other works on learned RA video compression, 
we do not restrict the training GoP size to 6 (the commonly used Vimeo-90K septuplet \cite{vimeo90k} dataset contains 7 frame sequences)
and develop training procedure for dataset with GoP size up to 32.

During training we utilize multi-stage strategy with ever increasing training GoP size: starting from 2 and up to 32. 
To achieve this with limited memory and to finish the training in a reasonable time, 
we propose the \emph{random path} sampling technique (\cref{fig:gop-b}) 
to train the model and reduce error propagation without compressing the entire GoP.
The idea is that on each training iteration, we randomly select one non-reference frame $x_t$ 
from the frames in the training sequence $\{x_1 ,..., x_{GoP}\}$
and compress only $x_t$ and all frames necessary to reconstruct it. This set of frames is called $path$. 
To account for such a change, the Rate-Distortion loss is weighted accordingly:

\begin{align}
    \label{eq:proposedloss}
    \mathcal{L}=R^I + \lambda D^I + \sum_{t\in\texttt{path}}{(R_t^m + R_t^{res}+\lambda D_t^B) \cdot 2^{level(t)}}
\end{align}
    
where $R_I$, $D_I$ -- rate and distortion of I-frame, $R_t^m$, $R_t^{res}$, $D_t^B$ -- rates and distortion of B-frame. 
Targeting comparison in YUV space, distortion here is weighted MSE:
    
\begin{align}
\label{eq:yuvdistortion}
D_t^B = \frac{8\cdot MSE_Y + c_t^{UV}\cdot(MSE_U + MSE_V)}{10}\cdot c_t
\end{align}

The random path sampling technique assumes that we compress only one B-frame from each temporal layer.
We multiply the rate and distortion of each compressed B-frame by the number
of B-frames in the same temporal layer, which equals $2^{level(t)}$, thus simulating the entire GoP sampling.

Some papers have already proposed to adjust the distortion weights in the loss at different levels of the hierarchical coding pyramid. 
\cite{BCANF} proposed to use 2 times lower distortion weight for non-reference B-frames. 
And \cite{HiDCVC} proposed to use different $\lambda$ values at different hierarchy levels. 
Unlike these two papers, we introduce additional coefficients $c_t$ and $c_t^{UV}$
which allow to flexibly balance distortion values between different hierarchy levels and between luma and chroma components. 
Specifying different $c_t$ coefficients for different $t$ simulates the dynamic QP approach of traditional codecs, 
forcing the model to spend more bits on far-reference frames and fewer bits on non-reference B-frames. 

We argue that extending the training data with long sequences is crucial for learned video compression, 
since it allows to include frames with motion distribution close to the test conditions. 
And while \cite{BCANFlike,HiDCVC} train only on 7-frame sequences, 
we believe that such a choice limits the generalization ability of the model and degrades the final results.

\section{Experiments}
\label{sec:experiments}

\subsection{Training}
For training, we use a proprietary dataset containing 33-frame sequences of different content.
The dataset does not overlap with the test sequences.
At a later stage, we also add sequences from the TVD \cite{TVD}.
The source sequences are resized to variable resolutions and randomly cropped to 256x256 and 512x512 patches.
We trained 4 models - one for each $\lambda = \{0.05, 0.015, 0.005, 0.001\}$, using the Adam \cite{adam} optimizer.

\begin{table}[h]
  \caption{Average weighted YUV-PSNR BD-rate results over VTM-17.0 (random access). Best results are highlighted in bold.}
  \label{tab:results_yuv}
  \centering
    \begin{tabular}{|c|c|c|c|c|c|c|c|}
    \hline
                     & UVG             & MCL-JCV         & JVET B          & JVET C          & JVET D          & JVET E          & Average         \\ \hline
    VTM-17.0 (RA)    & 0\%             & 0\%             & 0\%             & 0\%             & 0\%             & 0\%             & 0\%             \\ \hline
    VTM-17.0   (LDP) & 45.7\%          & 46.8\%          & 59.7\%          & 51.4\%          & 59.0\%          & 59.6\%          & 53.7\%          \\ \hline
    DCVC-FM          & 35.4\%          & 56.6\%          & 31.6\%          & 21.9\%          & -0.5\%          & 17.2\%          & 27.0\%          \\ \hline
    \textbf{Ours}    & \textbf{11.5\%} & \textbf{27.2\%} & \textbf{23.6\%} & \textbf{15.1\%} & \textbf{-7.7\%} & \textbf{-2.8\%} & \textbf{11.1\%} \\ \hline
    \end{tabular}
\end{table}

\begin{table}[h]
  \caption{Average VMAF BD-rate results over VTM-17.0 (random access). Best results are highlighted in bold.}
  \label{tab:results_vmaf}
  \centering
    \begin{tabular}{|c|c|c|c|c|c|c|c|}
    \hline
                     & UVG             & MCL-JCV        & JVET B          & JVET C           & JVET D           & JVET E           & Average          \\ \hline
    VTM-17.0 (RA)    & 0\%             & 0\%            & 0\%             & 0\%              & 0\%              & 0\%              & 0\%              \\ \hline
    VTM-17.0   (LDP) & 40.3\%          & 40.4\%         & 54.6\%          & 41.4\%           & 56.9\%           & 53.5\%           & 47.9\%           \\ \hline
    DCVC-FM          & 27.3\%          & 37.0\%         & 25.7\%          & 6.7\%            & -4.1\%           & -3.9\%           & 14.8\%           \\ \hline
    \textbf{Ours}    & \textbf{-1.3\%} & \textbf{1.4\%} & \textbf{-2.1\%} & \textbf{-14.6\%} & \textbf{-27.5\%} & \textbf{-33.4\%} & \textbf{-12.9\%} \\ \hline
    \end{tabular}
\end{table}

\begin{table}[h]
  \caption{Average YUV/Y/U/V-PSNR and VMAF BD-rates results for our model over VTM-17.0 (random access).}
  \label{tab:results_y_u_v}
  \centering
  \begin{tabular}{|c|c|c|c|c|c|c|c|}
  \hline
                        &        & JVET A1 & JVET A2 & JVET B  & JVET C  & JVET D  & JVET E  \\ \hline
  \multirow{2}{*}{YUV}  & w/o CA & 45.9\%  & 15.6\%  & 23.6\%  & 15.1\%  & -7.7\%  & -2.8\%  \\ \cline{2-8} 
                        & w/ CA  & 33.1\%  & 11.0\%  & 12.8\%  & 6.4\%   & -12.5\% & -       \\ \hline
  \multirow{2}{*}{VMAF} & w/o CA & 1.8\%   & -7.1\%  & -2.1\%  & -14.6\% & -27.5\% & -33.4\% \\ \cline{2-8} 
                        & w/ CA  & -7.5\%  & -13.2\% & -11.4\% & -19.1\% & -31.2\% & -       \\ \hline
  \multirow{2}{*}{Y}    & w/o CA & 25.2\%  & 7.0\%   & 21.5\%  & 12.4\%  & -8.9\%  & -6.6\%  \\ \cline{2-8} 
                        & w/ CA  & 14.4\%  & 2.8\%   & 10.5\%  & 4.0\%   & -13.4\% & -       \\ \hline
  \multirow{2}{*}{U}    & w/o CA & 160.3\% & 64.0\%  & 34.2\%  & 18.4\%  & -0.4\%  & 9.0\%   \\ \cline{2-8} 
                        & w/ CA  & 127.0\% & 58.3\%  & 22.3\%  & 9.8\%   & -6.6\%  & -       \\ \hline
  \multirow{2}{*}{V}    & w/o CA & 74.6\%  & 91.4\%  & 27.9\%  & 37.0\%  & -2.7\%  & 7.1\%   \\ \cline{2-8} 
                        & w/ CA  & 56.8\%  & 86.2\%  & 18.3\%  & 23.9\%  & -9.9\%  & -       \\ \hline
  \end{tabular}
\end{table}

\subsection{Testing in YUV Colorspace}
Here we present quantitative and qualitative results of the proposed solution. 
For a fair comparison, we follow the JVET common test conditions (CTC) \cite{jvetctc}. 
All tests on the JVET dataset are performed on the first 129 frames of the sequences.
We also tested the models on UVG \cite{uvg} and MCL-JCV \cite{mcljcv} datasets, which are commonly used in learned video compression.
Following other works, we use only the 1080p subset of UVG, consisting of 7 sequences.
MCL-JCV contains 30 1080p sequences.
In tests on the UVG dataset, we used the first 129 frames.
Some sequences from MCL-JCV are shorter than 129 frames, so all tests were performed on the first 97 frames for consistency.
Following CTC, we measure the model performance in YUV420 color space and provide Y, U, and V PSNRs in \cref{tab:results_y_u_v} and \cref{tab:pervideoresults}.
However, the weighted YUV-PSNR metric is more common for the latest E2E video compression work, so we provide such results in \cref{tab:results_yuv} (weights are equal to 6 for the Y component and 1 for the U and V components).
We also provide values for the VMAF\cite{VMAF} metric in \cref{tab:results_vmaf} as it is more correlated with human perception.

VVC reference software VTM-17.0 \cite{vtm} in random access configuration was chosen as the main anchor.
According to JVET CTC, the GoP size was set to 32 and the intra period was different for different frame rates. 
We also compare with VTM-17.0 in low-delay P configuration as the main baseline for most papers on learned video compression.
For this codec, the intra period was the same as for VTM RA.
Among the published works on end-to-end learned video coding, only a few also provide complete inference code and/or pretrained models.
In the YUV color space, we have tested DCVC-FM \cite{DCVCFM} as a state-of-the-art model in E2E LDP video compression.
For this model, the GoP size is the same as the intra period, which was set equal to the intra period values of VVC.
The same is true for our model: the GoP size is the same as the intra period set according to CTC.
In this way, all tested codecs always use the same number of intra frames (I-frames) and inter frames (P/B-frames).

As can be seen, the best results in terms of YUV-PSNR are achieved on JVET classes D and E.
With content adaptation enabled, our model also comes close to VTM 17 RA on JVET classes A2, B, C.
In terms of VMAF, our model outperforms traditional codecs on all JVET classes and on the UVG dataset.

\begin{figure}[tb]
  \centering
    \begin{subfigure}{0.49\linewidth}
        \includegraphics[width=\linewidth]{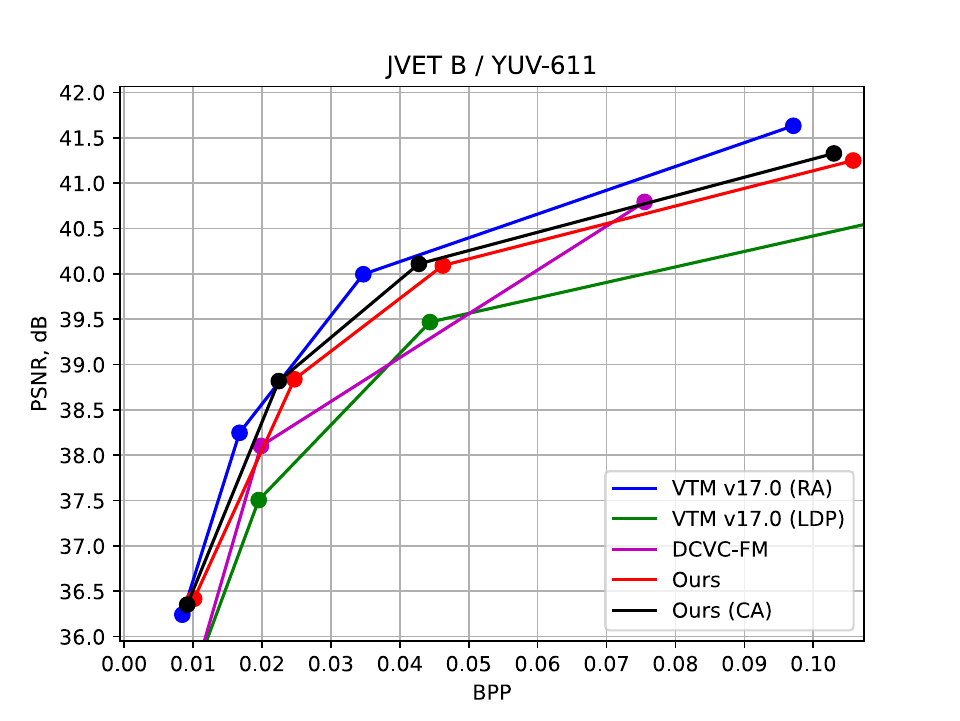}
        \caption{}
    \end{subfigure}
    \begin{subfigure}{0.49\linewidth}
        \includegraphics[width=\linewidth]{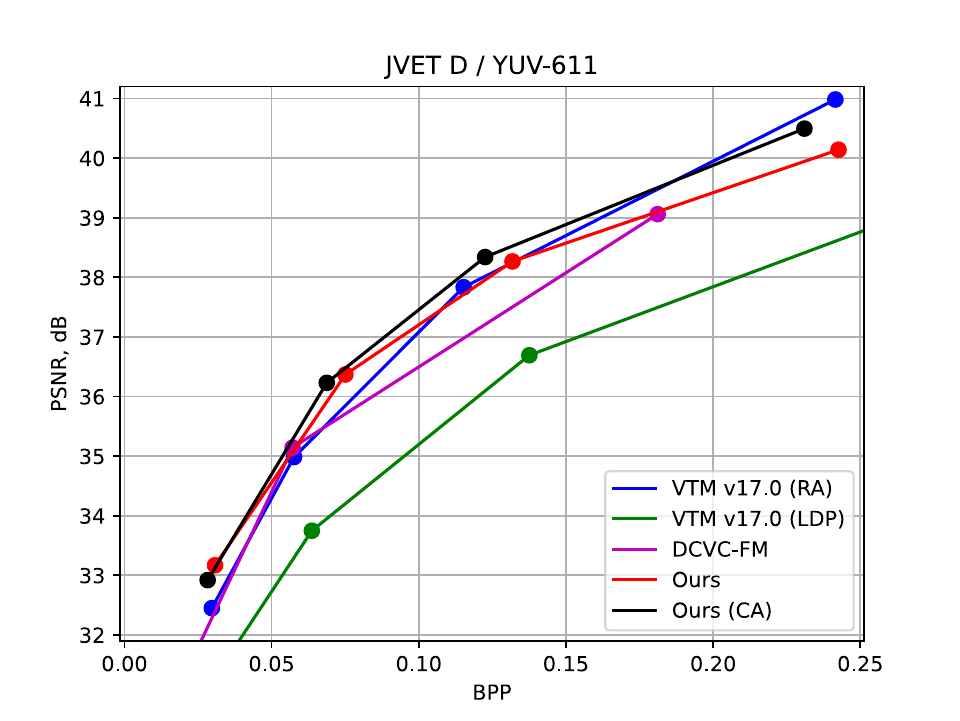}
        \caption{}
    \end{subfigure}
    \begin{subfigure}{0.49\linewidth}
        \includegraphics[width=\linewidth]{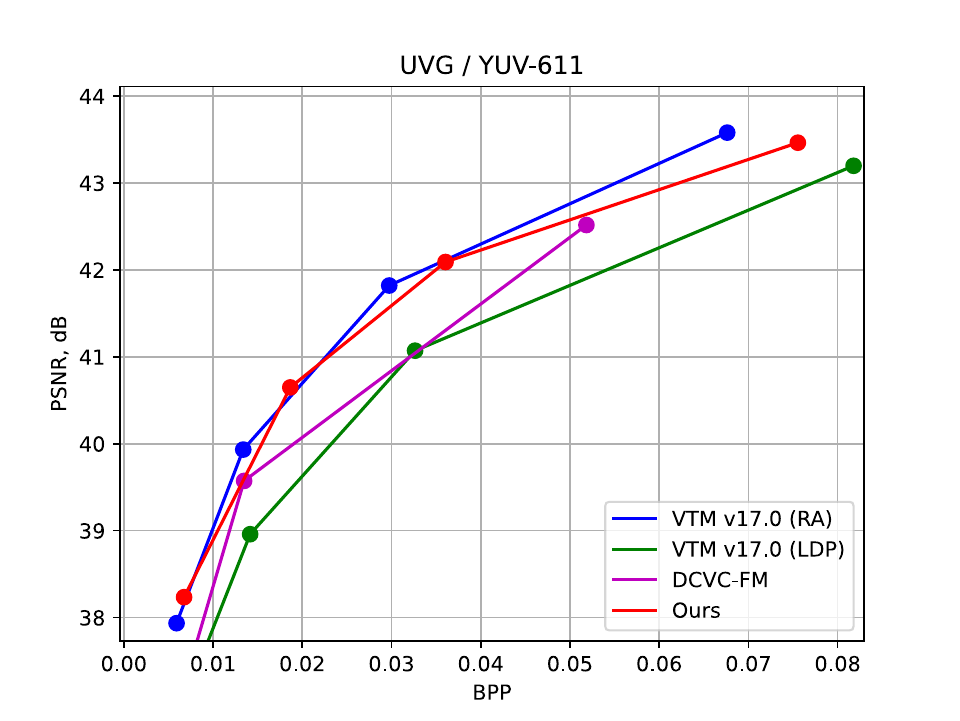}
        \caption{}
    \end{subfigure}
    \begin{subfigure}{0.49\linewidth}
        \includegraphics[width=\linewidth]{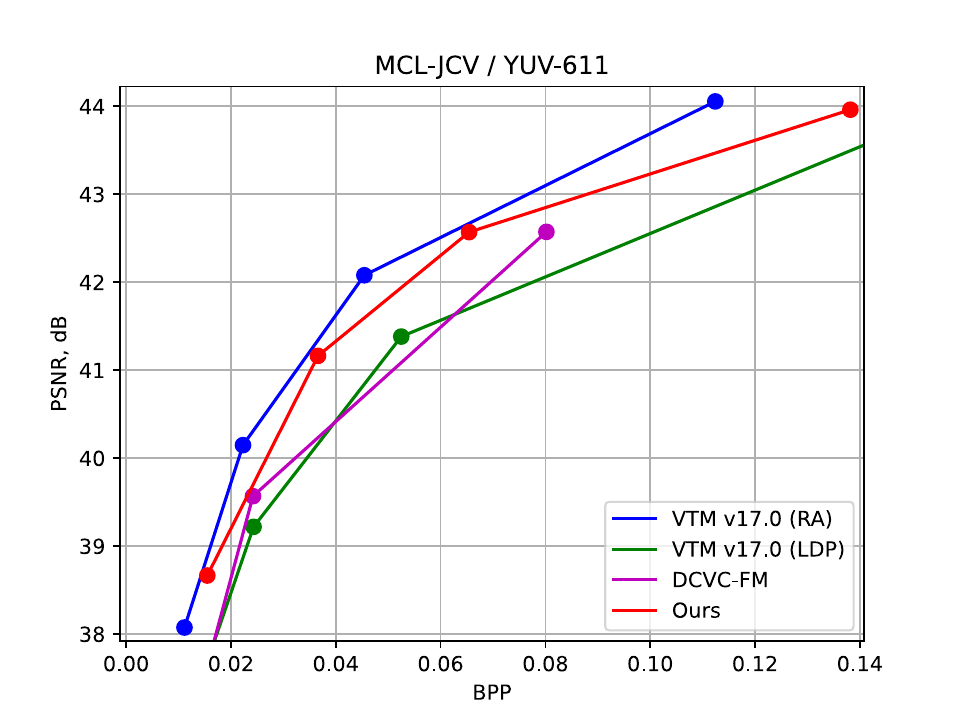}
        \caption{}
    \end{subfigure}
    \caption{YUV-PSNR Rate-Distortion curves for different methods.}
    \label{fig:plots_psnr}
\end{figure}

\begin{figure}[tb]
  \centering
    \begin{subfigure}{0.49\linewidth}
        \includegraphics[width=\linewidth]{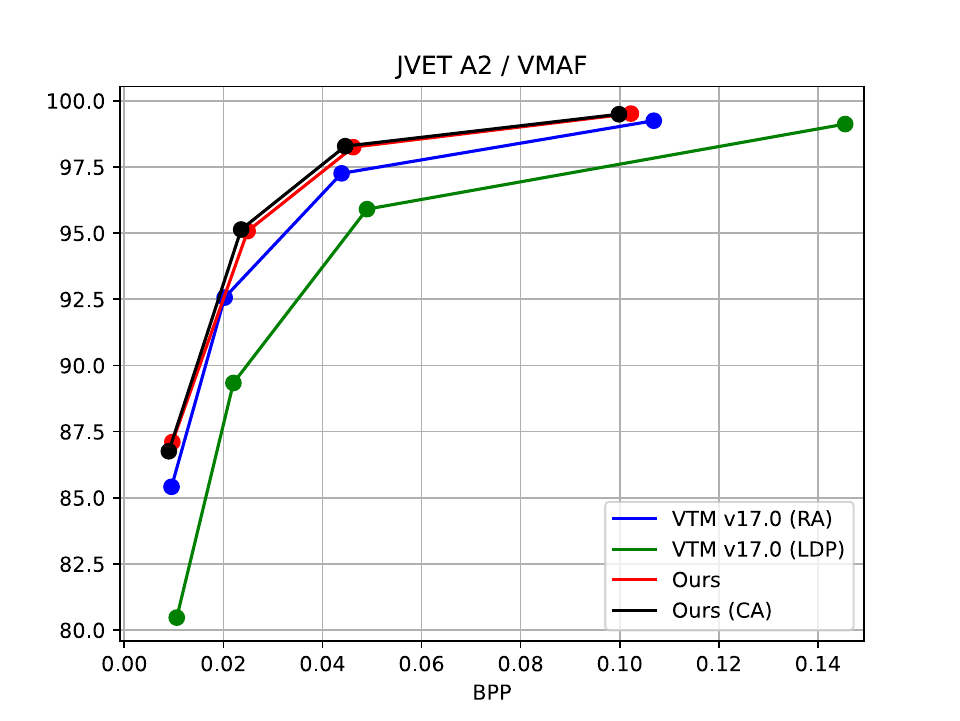}
        \caption{}
    \end{subfigure}
    \begin{subfigure}{0.49\linewidth}
        \includegraphics[width=\linewidth]{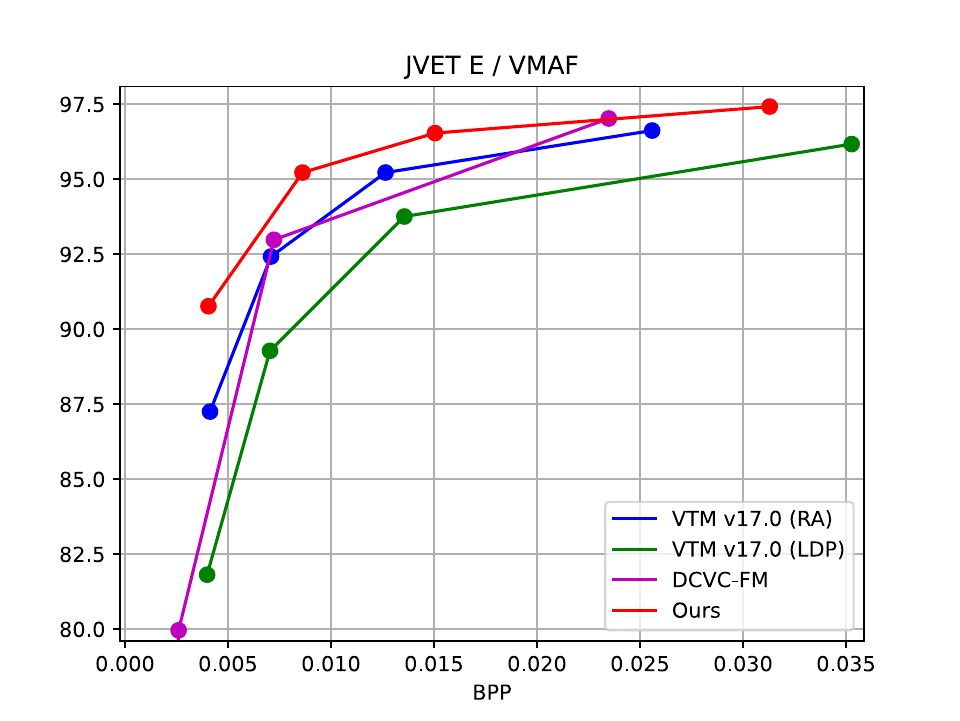}
        \caption{}
    \end{subfigure}
    \caption{VMAF Rate-Distortion curves for different methods.}
    \label{fig:plots_vmaf}
\end{figure}

Additionally, we provide YUV-PSNR and VMAF rate-distortion curves for different classes on \cref{fig:plots_psnr,fig:plots_vmaf}.

\subsection{Testing in RGB Colorspace}
Recent papers on end-to-end RA video compression propose solutions trained with RGB-PSNR as the target metric.
In order to make a fair comparison with them, the following tests were performed in RGB color space. 
To obtain correct inputs for such models, 
YUV 4:2:0 input is converted to YUV 4:4:4 by UV upsampling and then converted to RGB using BT.709 conversion matrix. 
For RGB-PSNR evaluation, the output of the proposed model is converted to RGB in the same way.
\cite{Cetin}, \cite{BCANFlike} and \cite{BCANF} were selected as competitors.
In contrast to the previous test, here the GoP size was reduced to 16 due to hardcoded restrictions of \cite{Cetin}, but number of frames is still equal to 129.
Despite training on YUV-specific loss and quality degradation due to color space conversion, our model is more effective as can be seen from \cref{tab:rgb_table}.

\begin{table}[h]
  \caption{RGB-PSNR BD-rate in GoP 16 setting, where anchor is our model.}
  \label{tab:rgb_table}
  \centering
    \begin{tabular}{|c|c|c|c|c|}
    \hline
                      & JVET B          & JVET C          & JVET D          & JVET E          \\ \hline
    Cetin \etal (2022) \cite{Cetin}    & 49.3\%             & 65.1\%             & 80.9\%             & 26.5\%                        \\ \hline
    TLZMC \cite{BCANFlike} & 174.6\%          & 143.8\%          & 110.1\%          & 109.1\%                   \\ \hline
    B-CANF \cite{BCANF}         & 5.3\%          & 20.3\%          & 11.5\%          & 8.8\%                   \\ \hline
    
    \end{tabular}
\end{table}

\begin{table}[h]
  \caption{Models complexity comparison.}
  \label{tab:complexity}
  \centering
  \begin{tabular}{|c|c|c|c|c|}
  \hline
          & MACs  & \# of params & Enc. time & Dec. time \\ \hline
  Ours    & 3888G & 35.14M       & 2.5s     & 0.94s     \\ \hline
  DCVC-DC & 2642G & 19.78M       & 1.1s      & 0.74s     \\ \hline
  \end{tabular}
\end{table}

\subsection{Complexity}

In the \cref{tab:complexity} we provide a complexity comparison. All values were measured on a 1080x1920 input frame on an NVidia V100 GPU.  

The finetuning of our model takes 342.8 s (with 150 optimization steps) under the same test conditions.
These values include only P-frames in the case of DCVC-DC and only B-frames in our case (I-frames are not included).

\section{Ablation}
\label{sec:ablation}
      
To determine the influence of each component on the proposed method, several experiments were performed, summarized in \cref{tab:ablationtable}.
The last column shows the weighted YUV-PSNR BD-rate of our best model, averaged over JVET classes A1, A2, B, C, D, E.
Check marks in the cells indicate that the corresponding tools are enabled.

ND stands for New Dataset, which replaced the commonly used Vimeo90k.

LG means training with Large Gop sizes (up to 32). 
The baseline model was trained with a maximum GoP size of 8. 
As can be seen, training with larger GoP sizes provides a huge gain on inference under real conditions.

HGU is an implementation of level-adaptive latent scaling. 
We add these blocks on late stages with GoP size $\geq 8$, initializing the scaling values with ones. 

L means the introduction of rebalancing loss coefficients $c_t \not = 1 , c_t ^{UV} \not = 1$.
We have configured these parameters to significantly reduce the quality loss in the chroma component.
This leads to a slight decrease in Y-PSNR: the ratio of luma loss to chroma gain is about 1:15, which seems reasonable to us.

CA stands for the Content Adaptation.
This is one of the main advantages of the learned video compression approach.
As mentioned in \cref{sec:contad}, we only tune the encoder-side modules.
This requires no additional bitrate and the only cost is encoding time.
The tuning results in a BD-rate gain of 0 to 22\% depending on the input video.

\begin{table}[]
    \caption{Ablation study.}
    \label{tab:ablationtable}
    \centering
      \begin{tabular}{ccccc|c}
      ND & LG & HGU & L & CA & YUV BD-Rate \\ \hline
         &    &     &   &    & 48.0\%      \\ \hline
      \checkmark  &    &     &   &    & 42.6\%      \\ \hline
      \checkmark  & \checkmark  &     &   &    & 17.5\%      \\ \hline
      \checkmark  & \checkmark  & \checkmark   &   &    & 15.3\%      \\ \hline
      \checkmark  & \checkmark  & \checkmark   & \checkmark &    & 7.0\%       \\ \hline
      \checkmark  & \checkmark  & \checkmark   & \checkmark & \checkmark  & 0.0\%      
      \end{tabular}
  \end{table}

\section{Conclusion}

In this paper we present an efficient learned video compression solution for the RA scenario.
In addition, a novel training approach has been developed that takes into account the specificities of the hierarchical coding scheme.

The presented ablation study shows the importance of training on large GoP sizes, which leads to better generalization.
We also propose several RA-specific tools such as Hierarchical Gain Unit and special hierarchy-dependent loss.
Finally, an RA-specific content adaptation procedure was developed to use all advantages of the learned video compression.

Among end-to-end RA codecs, we are the first to our knowledge to compare fairly with VTM under common test conditions in the YUV colorspace.
Our codec shows promising and sometimes superior results in terms of YUV-PSNR BD rate and outperforms traditional codecs in terms of VMAF BD-rate.

Conducted experiments showed the critical role of the motion estimation (ME) 
and motion compensation (MC) for the B-frame performance due to large reference distances.
To further investigate this problem, we will continue to develop of the B-frame architecture, especially the ME and MC modules.
We also plan to continue experiments with the new training data and training strategy optimization.
Perceptual quality improvement is also an area of interest. 
Since PSNR does not correlate well with human perception, we want to explore other training objectives such as LPIPS.

\bibliographystyle{splncs04}
\bibliography{main}

\appendix
\section{Appendices}
\label{sec:appendices}

\subsection{I-frame}

As an I-frame model we use ELIC \cite{elic}. 
The model is first pretrained for the image compression task 
using the same dataset, optimizer, and loss function as described in \cref{sec:training}.
Then, it is incorporated into the B-frame training procedure for I-frame compression. 
In the early stages of training, the weights of the model are frozen. 
But in later stages, it is trained together with the B-frame model.

\subsection{Testing}

\paragraph{Intraperiod.}

As stated in the paper, in order to satisfy the common test conditions (CTC), the intra period is chosen based on the FPS of the video sequence.
Among the tested JVET classes (A1, A2, B, C, D and E) there are only 3 videos with 30 FPS: Campfire (class A1), RaceHorses (class C) and RaceHorses (class D).
For them the intra period is set to 32, while for other sequences it's equal to 64.
All UVG sequences are tested with intra period 64 and all MCL-JCV sequences are tested with intra period 32.
For learned video codecs in low-delay mode, the last frame of each sequence is coded as an I-frame. 
In this way, we ensure an equal number of I-frames among all tested methods.

\paragraph{Metrics.}

The reported values of PSNR over sequence are calculated as the average of PSNR over all frames.
The official implementation~\footnote{\url{https://github.com/Netflix/vmaf/blob/master/resource/doc/python.md}} was used for the VMAF calculation: 
configuration v0.6.1\_4k for classes A1, A2 and configuration v0.6.1 for other classes.
Following standard practices, the BD rate per video class is defined as the average of the BD rates over all sequences in it.

\paragraph{Settings of Codecs.}

To test VTM 17.0 in random access scenario, we take $QP$ from $\{22, 27, 32, 37\}$, which results in the following command:

\begin{itemize}
   \item {\bf VTM}\par EncoderAppStatic  \par
   -c encoder\_randomaccess\_vtm.cfg \par
   -i \{{\em path to yuv video}\} \par
   -wdt \{{\em width}\} \par
   -hgt \{{\em height}\} \par
   -{}-InputBitDepth=\{{\em bit depth}\} \par
   -fr \{{\em FPS}\} \par
   -f 129 \par
   -q \{{\em QP}\} \par
   -{}-IntraPeriod=\{{\em intraperiod}\} \par
   -{}-BitstreamFile=\{{\em bitstream file name}\}
\end{itemize}

To test VTM 17.0 LDP, the same command was used as for VTM RA
but with the options {\em -c encoder\_lowdelay\_P\_vtm.cfg} and {\em -{-}DecodingRefreshType=2}.

The next few sentences provide details on testing the learned video codecs.

Since DCVC-FM doesn't support 10-bit input by default, we modify it by scaling the input and output of the model according to the input bit depth.
This allows us to correctly handle class B videos (MarketPlace, RitualDance). The tests were run using scripts provided by the authors.

Since TLZMC~\footnote{\url{https://github.com/nycu-clab/tlzmc-cvpr}} provides checkpoints for 3 models ({\em plus}, {\em star}, {\em dstar}), 
we tested all of them and chose the best one for comparison.
Although this model supports different GoP sizes (even up to 64), we only provide a test with GoP 16.

That's because another B-frame model was chosen for comparison. 
(Cetin \etal (2022)~\footnote{\url{https://github.com/KUIS-AI-Tekalp-Research-Group/video-compression/tree/master/Flex-Rate-Hier-Bidir-Video-Compression}}) 
limits the GoP size to 16, so the inference configuration is only applicable to this case.

\paragraph{Visual Comparison.}

\begin{figure}[tb]
  \centering
    \begin{subfigure}{\linewidth}
        \includegraphics[width=\linewidth]{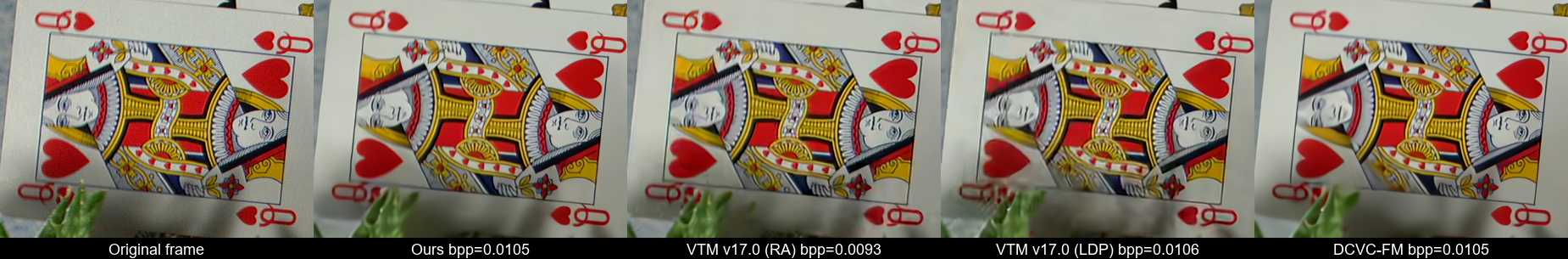}
    \end{subfigure}
    \hfill
    \centering
    \begin{subfigure}{\linewidth}
        \includegraphics[width=\linewidth]{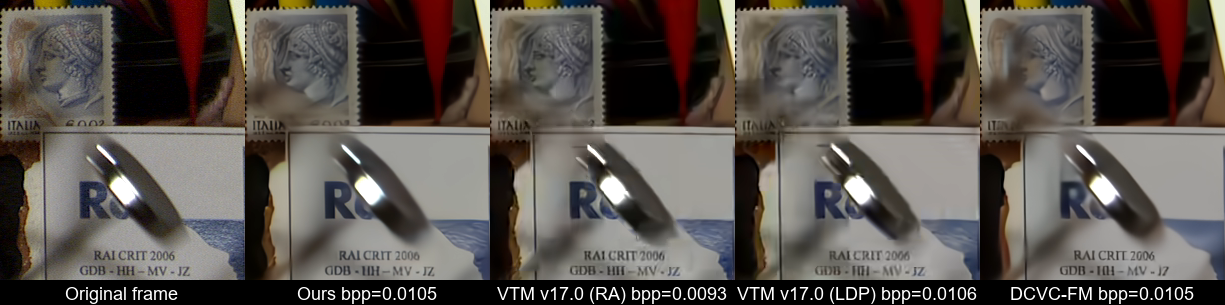}
    \end{subfigure}
    \hfill
    \centering
    \begin{subfigure}{\linewidth}
        \includegraphics[width=\linewidth]{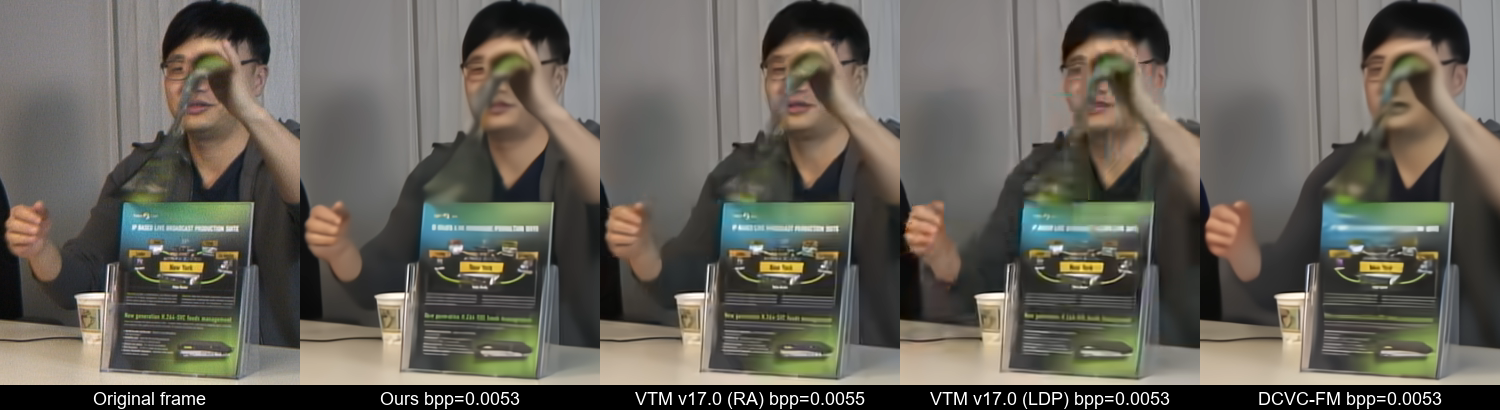}
    \end{subfigure}
  \caption{Comparison of different methods.}
  \label{fig:compares}
  \end{figure}

\Cref{fig:compares} shows the visual comparison of our codec with other solutions on sequences of different JVET classes. It can be seen that our method provides better image quality without blocking artifacts and with better clarity.

  \begin{table}[b]
    \caption{BD-rate results on JVET CTC over VTM-17.0 (random access).}
    \label{tab:pervideoresults}
    \centering
    \begin{tabular}{|c|c|c|c|c|}
    \hline
                              &                 & Y       & U       & V       \\ \hline
    \multirow{3}{*}{Class A1} & Tango2          & 10.2\%  & 39.5\%  & 47.3\%  \\ \cline{2-5} 
                              & FoodMarket4     & 47.5\%  & 98.0\%  & 107.6\% \\ \cline{2-5} 
                              & Campfire        & 17.9\%  & 343.4\% & 68.9\%  \\ \hline
    \multirow{3}{*}{Class A2} & CatRobot        & 2.6\%   & 20.8\%  & 36.2\%  \\ \cline{2-5} 
                              & DaylightRoad2   & 25.3\%  & 30.1\%  & 59.3\%  \\ \cline{2-5} 
                              & ParkRunning3    & -6.8\%  & 140.9\% & 178.6\% \\ \hline
    \multirow{5}{*}{Class B}  & MarketPlace     & 6.0\%   & 8.1\%   & 2.6\%   \\ \cline{2-5} 
                              & RitualDance     & 0.2\%   & 33.6\%  & 21.3\%  \\ \cline{2-5} 
                              & Cactus          & -1.6\%  & 3.3\%   & 14.0\%  \\ \cline{2-5} 
                              & BasketballDrive & 11.2\%  & 33.7\%  & 62.7\%  \\ \cline{2-5} 
                              & BQTerrace       & 91.4\%  & 92.4\%  & 38.9\%  \\ \hline
    \multirow{4}{*}{Class C}  & BasketballDrill & -7.5\%  & 11.7\%  & 32.8\%  \\ \cline{2-5} 
                              & BQMall          & 14.4\%  & 15.7\%  & 72.6\%  \\ \cline{2-5} 
                              & PartyScene      & -0.7\%  & 13.8\%  & 12.5\%  \\ \cline{2-5} 
                              & RaceHorses      & 43.6\%  & 32.4\%  & 30.0\%  \\ \hline
    \multirow{4}{*}{Class D}  & BasketballPass  & -11.8\% & -6.9\%  & 2.2\%   \\ \cline{2-5} 
                              & BlowingBubbles  & -9.0\%  & 11.2\%  & -3.1\%  \\ \cline{2-5} 
                              & BQSquare        & -14.9\% & -17.8\% & -25.4\% \\ \cline{2-5} 
                              & RaceHorses      & 0.0\%   & 12.0\%  & 15.3\%  \\ \hline
    \multirow{3}{*}{ClassE}   & FourPeople      & -12.5\% & -7.5\%  & -7.1\%  \\ \cline{2-5} 
                              & Johnny          & 0.9\%   & 23.7\%  & 14.0\%  \\ \cline{2-5} 
                              & KristenAndSara  & -8.2\%  & 10.7\%  & 14.4\%  \\ \hline
    \end{tabular}
\end{table}

\end{document}